\documentclass{article}
\usepackage{spconf,amsmath,graphicx}
\usepackage{url}            

\graphicspath{{./}}

\title{Virtualization of tissue staining in digital pathology using an unsupervised deep learning approach}
 \name{Amal Lahiani$^{\star \dagger}$ \! Jacob Gildenblat$^{\P}$ \! Irina Klaman$^{\star}$ \! Shadi Albarqouni$^{\dagger}$ \! Nassir Navab$^{\dagger}$ \! Eldad Klaiman$^{\star}$}

 \address{$^{\star}$ {\normalsize Pathology and Tissue Analytics, Pharma Research and Early Development, Roche Innovation Center Munich} \\
     $^{\dagger}$ {\normalsize Computer Aided Medical Procedures, Technische Universit\"at M\"unchen} \\
     $^{\P}$ {\normalsize DeePathology Ltd.}}
\begin{document}
%
\maketitle
\footnote{This work has been submitted to the IEEE for possible publication. Copyright may be transferred without notice, after which this version may no longer be accessible.}
\begin{abstract}
Histopathological evaluation of tissue samples is a key practice in patient diagnosis and drug development, especially in oncology. Historically, Hematoxylin and Eosin (H\&E) has been used by pathologists as a gold standard staining. However, in many cases, various target specific stains, including immunohistochemistry (IHC), are needed in order to highlight specific structures in the tissue. As tissue is scarce and staining procedures are tedious, it would be beneficial to generate images of stained tissue virtually. Virtual staining could also generate in-silico multiplexing of different stains on the same tissue segment. In this paper, we present a sample application that generates FAP-CK virtual IHC images from Ki67-CD8 real IHC images using an unsupervised deep learning approach based on CycleGAN. We also propose a method to deal with tiling artifacts caused by normalization layers and we validate our approach by comparing the results of tissue analysis algorithms for virtual and real images.
\end{abstract}
\section{Introduction}
\label{sec:intro}

In the field of pathology, staining types determine which parts or targets in the tissue are highlighted with specific colors. Tissue staining materials and procedures can be time consuming, expensive, and typically require special expertise. These limitations usually reduce the number of examinations and stainings performed on a sample. This can limit clinicians' ability to obtain all relevant information from a patient biopsy. In many cases information exists in the stained slide image about targets and objects not specifically targeted by the stain. For example, pathologists have the ability to identify lymphocytes in a Hematoxylin and Eosin (H\&E) image \cite{1} even without directly staining them for lymphocyte specific markers. This fact motivated the research in the direction of generating virtually stained slides from other modalities \cite{3, 4, 6, 8, 9}. Recently, supervised deep learning based methods have been applied in the task of virtual staining generation  \cite{10, 23, 11, 12}. As supervised training methods are based on coupled pairs of aligned images, all the aforementioned methods require additional accurate registration steps between dataset image pairs.

In this work we propose to virtually generate FAP-CK stained slide images from Ki67-CD8 stained slide images. These input and output stainings were chosen for several reasons. First, information about tumor characteristics in FAP-CK could be encoded in the form of proliferation and tumor infiltrating lymphocytes in Ki67-CD8. Furthermore, Ki67-CD8 is one of the classical Immunohistochemistry (IHC) stainings used in histopathology while FAP-CK is a new duplex IHC protocol allowing to characterize tumor and to advance research in the direction of drug development. Additionally, generating virtual FAP-CK stained slide images from Ki67-CD8 allows the creation of a virtual multiplexed brightfield image, i.e. having 4 target stains on the same whole slide coordinate system, which is technically challenging using classical staining methods. In this paper, we present an unsupervised deep learning method based on Cycle-Consistent Adversarial Networks (CycleGAN) \cite{14}. This allows avoiding the slide registration process for training datasets and facilitates dealing with variability present in sets of slide images due to different lab protocols, scanners and experiment conditions. We further present a method aimed at reducing the tiling artifact caused by tile-wise processing of large images, a common problem in image style transfer encountered when high resolution testing images can not fit into memory \cite{15}. Finally, we validate the results of our method by comparing quantification of tumor cells and FAP in virtual slides with a real stained slide taken from the same tissue block.

\section{Methodology}
\label{sec:format}

\subsection{Dataset}

\begin{figure}[t]

\begin{minipage}[b]{.48\linewidth}
  \centering
  \centerline{\includegraphics[width=.7\linewidth]{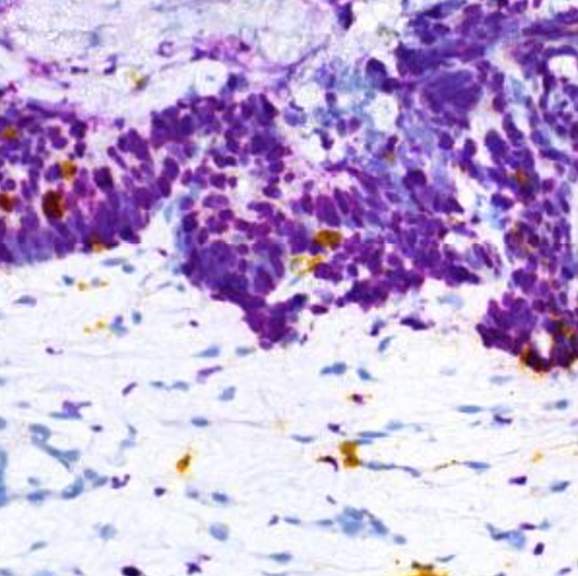}}
  \centerline{(a) Ki67-CD8 tile.}\medskip
\end{minipage}
\hfill
\begin{minipage}[b]{0.48\linewidth}
  \centering
  \centerline{\includegraphics[width=.7\linewidth]{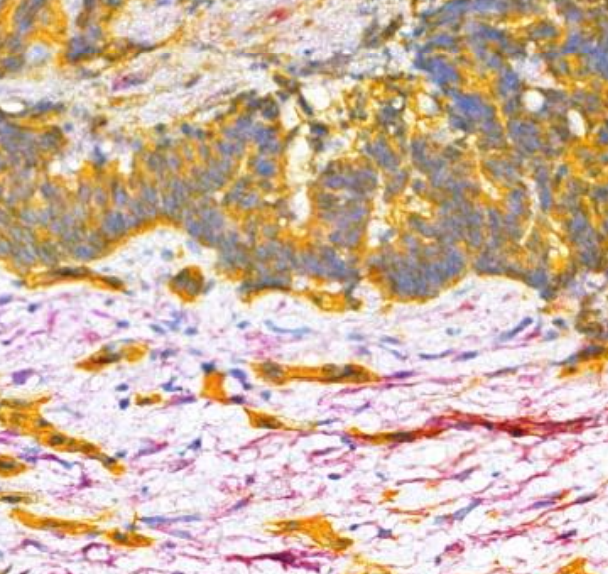}}
  \centerline{(b) FAP-CK tile.}\medskip
\end{minipage}
\vspace{-0.3cm}

\caption{Ki67-CD8 and FAP-CK example tiles.}
\label{examples_real}
\end{figure}

We selected a subset of whole slide images (WSI) of Colorectal Carcinoma metastases in liver tissue from biopsy and surgical specimen from Roche Pathology image database. All the slides were chosen following a review of tissue, staining and image quality. The training dataset includes 20 images: 10 from Ki67-CD8 stained slides and 10 from FAP-CK stained slides,  each from different patients. As high resolution whole slide histology images contain billions of pixels with 20x magnification and hardware memory is limited, it is necessary to tile the image into smaller segments for analysis and when possible use lower magnification. For these reasons, slides were tiled into overlapping $512 \times 512$ images at 10x magnification. The reduced magnification allows to have enough contextual information in the input which is needed in order to learn a meaningful feature set in the model while at the same time facilitates dealing with the computational memory limits \cite{17}. The tiling yielded 17025 tiles from Ki67-CD8 slides and 17812 tiles from FAP-CK slides. In order to test the performance of our method, we selected 10 pairs of Ki67-CD8/FAP-CK slides from the same tissue block. Fig. \ref{examples_real} shows examples of Ki67-CD8 and FAP-CK tiles.

\subsection{Network architecture and inference}
In order to map images from one staining to the other, we used a CycleGAN \cite{14} based method which allows to learn transforms between domain spaces without the need to use paired registered image datasets. As generator networks we used ResNet architectures \cite{16} with 11 residual blocks. The ResNet architecture used as a generator has a receptive field of $207 \times 207$ pixels, roughly corresponding to $190 \times 190$ microns on the $10x$ magnification image of the tissue. This receptive field size allows enough contextual information from the surroundings of the target pixel in the input to find meaningful histology features for the prediction of the virtual stain on the output image. Due to the CycleGAN memory burden, we were not able to fit more than one $512 \times 512$ RGB image per batch during training per GPU (Nvidia P100). In order to distribute the training on multiple GPUs and accelerate the learning process, we implemented the stochastic synchronous ADAM algorithm \cite{18} and used the pytorch distributed computing library which allowed us to use 12 GPUs concurrently on the Roche Pharma HPC cluster in Penzberg.

\begin{figure}[t]

  \centering
  \centerline{\includegraphics[width=1.0\linewidth]{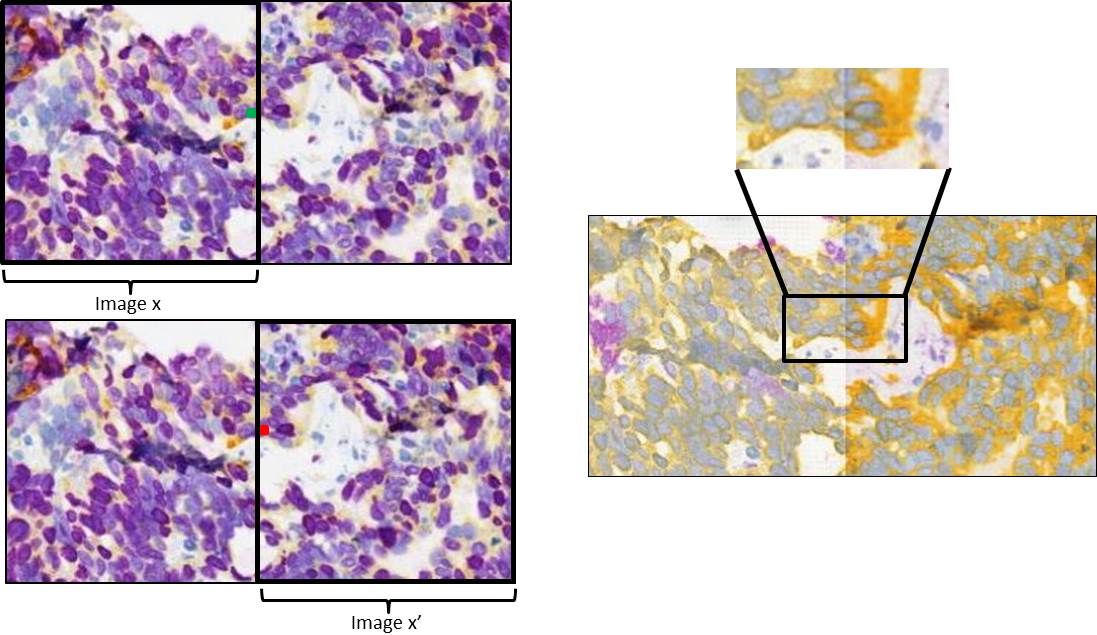}}
  
\caption{(right) Tiling effect in adjacent output tiles. (left) Image $x$ and $x^{'}$ correspond to 2 adjacent input tiles from a whole slide image. The green and red circles correspond to 2 adjacent pixels belonging to the same cell nucleus but to different input tiles.}
\label{tiling_artifact}
\end{figure}

In order to overcome the memory bound hardware limitations, inference of the trained network on the testing slides was also done tilewise. The tile output is then merged back in order to obtain a virtual whole slide image. This inference workflow yielded whole slide images containing tiling artifacts. As style transfer networks give better and more realistic results when instance normalization layers are used \cite{22}, these layers were introduced in the CycleGAN architecture. Instance normalization layers are applied at test time as well, making the value of any output pixel depends not only on the network parameters and the receptive field area in the input but also on the statics of the input tile image. Let's assume that:
\vspace{-0.3cm}
\begin{center}
$y =  g(x)$,
\end{center}
\vspace{-0.2cm}

where $x$, $y$ and $g$ correspond to an input tensor, an output tensor and an instance normalization layer respectively. Let $x \in {\rm I\!R}^{T \times C \times W \times H}$ be a tensor containing a batch of $T$ images and $x_{tijk}$ the $tijk^{th}$ element, where $j$ and $k$ correspond to spatial dimensions, $i$ corresponds to the feature channel and $t$ corresponds to the batch index. In this case, $g$ can be expressed as:

\vspace{-0.3cm}
\begin{center}
$y_{tijk} = g(x_{tijk}) = \frac{x_{tijk} - \mu_{ti}}{\sqrt{{\sigma_{ti}}^2 + \varepsilon}} $,
\end{center}
\vspace{-0.2cm}

where $\mu_{ti}$ and ${\sigma_{ti}}^2$ are the mean and variance of the input tile. If we consider 2 adjacent pixels $x_{tijk}$ and ${x^{'}}_{tijk}$ with very similar values on the edges of two adjacent input tiles $x$ and $x^{'}$ having very different statistics, the instance normalization functions $g$ and $g^{'}$ applied to these two pixels will be completely different. This results in a tiling artifact in the generated whole slide image as adjacent output tiles might have significantly different pixel values on their borders (Fig. \ref{tiling_artifact}).

\begin{figure}[t]

\begin{minipage}[t]{.48\linewidth}
  \centering
  \vspace{0pt}
  \centerline{\includegraphics[width=4.0cm]{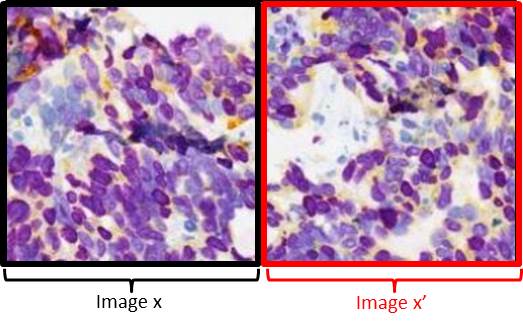}}
  \vspace{0.32cm}
  \centerline{(a)}\medskip
\end{minipage}
\hfill
\begin{minipage}[t]{0.48\linewidth}
  \centering
  \vspace{0pt}
  \centerline{\includegraphics[width=4.0cm]{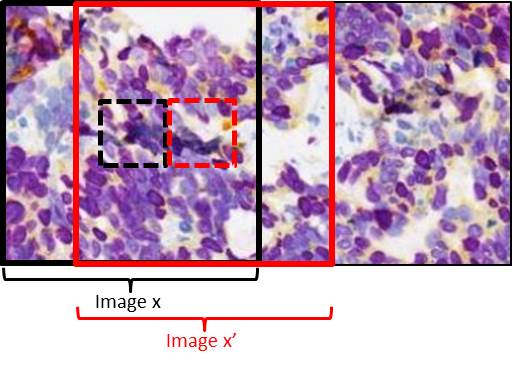}}
  \centerline{(b)}\medskip
\end{minipage}
\vspace{-0.3cm}
\caption{Image (a) corresponds to the inference performed on 2 adjacent slides using the classical method. 
Image (b) corresponds to the new inference approach. The solid and dotted line squares correspond to the sliding window considered and the effective tile used for inference respectively.
}
\label{deploy_methods}
\end{figure}

\begin{figure}[t]

\begin{minipage}[b]{.48\linewidth}
  \centering
  \centerline{\includegraphics[width=1.0\linewidth]{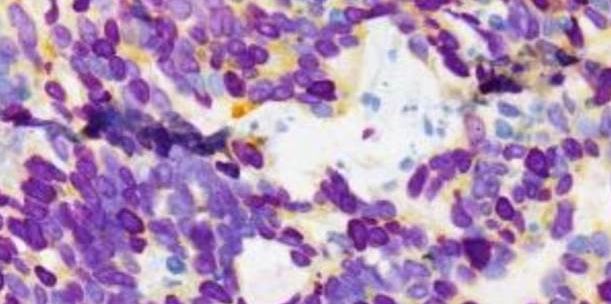}}
  \centerline{(a)}\medskip
\end{minipage}
\hfill
\begin{minipage}[b]{.48\linewidth}
  \centering
  \centerline{\includegraphics[width=1.0\linewidth]{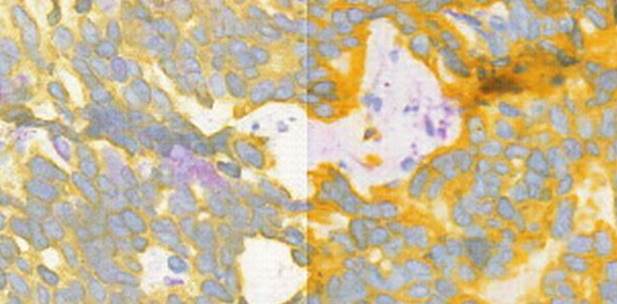}}
  \centerline{(b)}\medskip
\end{minipage}
\begin{minipage}[b]{.48\linewidth}
  \centering
  \centerline{\includegraphics[width=1.0\linewidth]{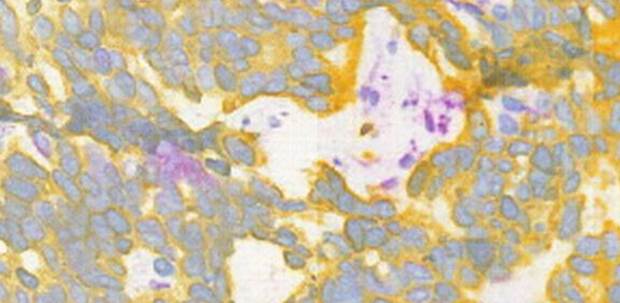}}
  \centerline{(c)}\medskip
\end{minipage}
\hfill
\begin{minipage}[b]{.48\linewidth}
  \centering
  \centerline{\includegraphics[width=1.0\linewidth]{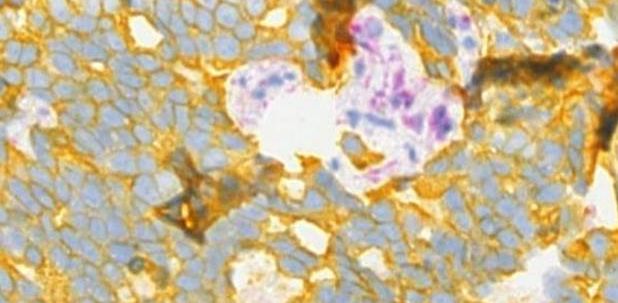}}
  \centerline{(d)}\medskip
\end{minipage}
\vspace{-0.3cm}

\caption{Effect of the new inference approach. (a) corresponds to 2 adjacent tiles from the input Ki67-CD8 image. (b) and (c) show to the corresponding image area in the output virtual FAP-CK image using the classical inference method and the proposed solution respectively. (d) shows a corresponding area from a real FAP-CK image from the same tissue block.}
\label{inference_effect}
\end{figure}

With instance normalization it is possible to use the same running mean and variance values for all the tiles at inference time in a way similar to inference with batch normalization. This approach indeed yielded output images without the tiling artifact. However, the resulting output at inference was of lower quality and had very faint colors. This effect can be explained by the fact that the training datasets are quite variable and containing both tissue and background, this makes the running mean and variance locally irrelevant. We propose a solution to the tiling artifact problem by using overlapping tiles during inference. Our solution is based on using a smaller input size of $128 \times 128$ instead of $512 \times 512$ and on using a sliding window for the instance normalization function statistics, allowing to have a smooth transition in the statistic values when deploying on 2 adjacent slides (Fig. \ref{deploy_methods}). Using this approach, we manage to substantially reduce the tiling artifact (Fig. \ref{inference_effect}).

\section{Results and validation}

\begin{figure}[t]

\begin{minipage}[b]{.3\linewidth}
  \centering
  \centerline{\includegraphics[width=1.0\linewidth]{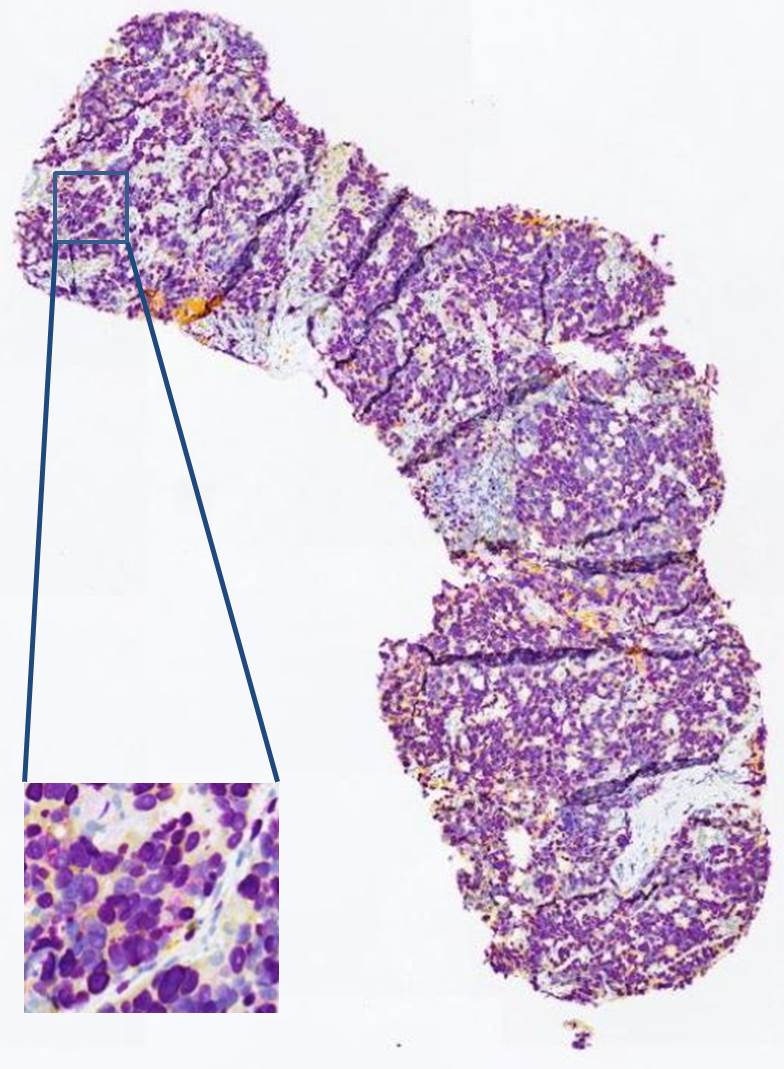}}
  \centerline{(a)}\medskip
\end{minipage}
\hfill
\begin{minipage}[b]{.3\linewidth}
  \centering
  \centerline{\includegraphics[width=1.0\linewidth]{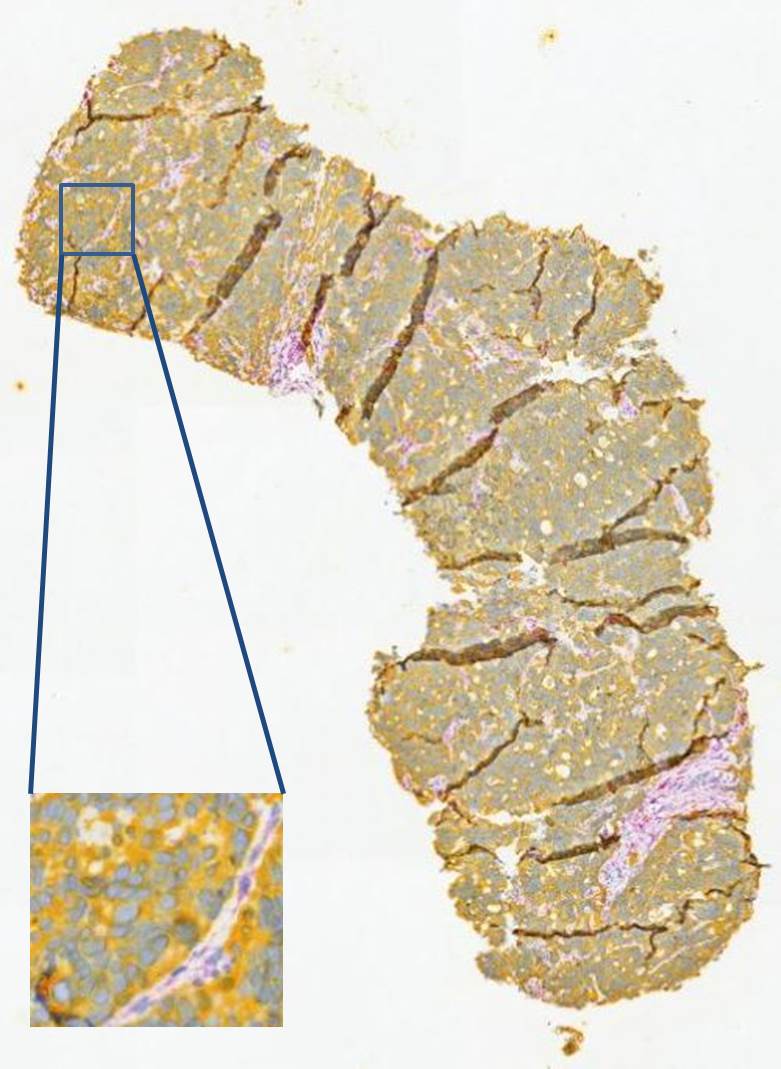}}
  \centerline{(b)}\medskip
\end{minipage}
\hfill
\begin{minipage}[b]{.3\linewidth}
  \centering
  \centerline{\includegraphics[width=1.0\linewidth]{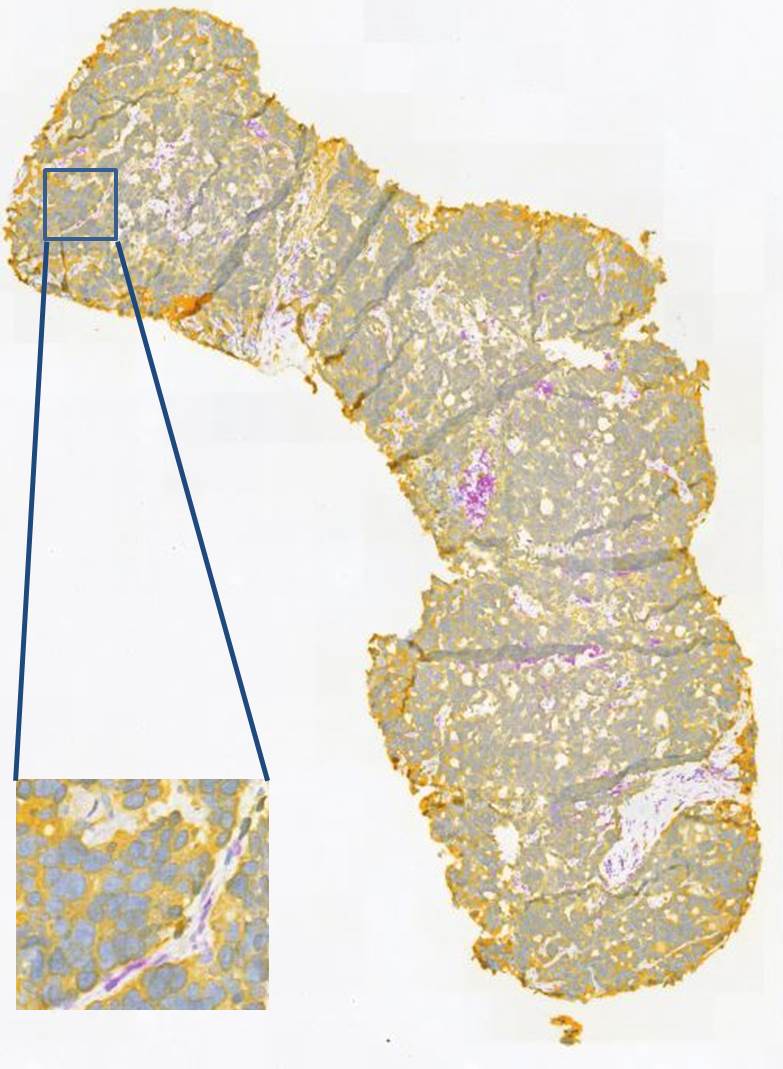}}
  \centerline{(c)}\medskip
\end{minipage}

%
\vspace{-0.3cm}
\caption{(a), (b) and (c) correspond to an input Ki67-CD8 image, the image of a real stained FAP-CK slide from the same tissue block and the virtual FAP-CK slide image.}
\label{results}
\end{figure}

Visual assessment of the virtual generated images shows that the results are visually similar to the real staining of a slide from the same tissue block (Fig. \ref{results}).

\begin{figure}[t]

\begin{minipage}[b]{.48\linewidth}
  \centering
  \centerline{\includegraphics[width=0.6\linewidth]{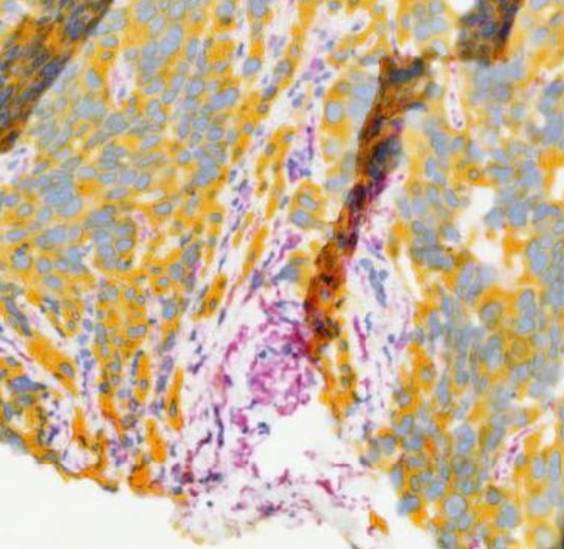}}
  \centerline{(a) real FAP-CK tile.}\medskip
\end{minipage}
\hfill
\begin{minipage}[b]{0.48\linewidth}
  \centering
  \centerline{\includegraphics[width=0.6\linewidth]{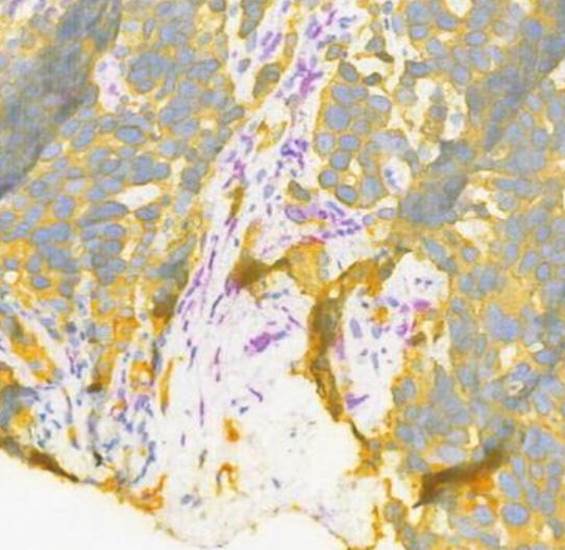}}
  \centerline{(b) Virtual FAP-CK tile.}\medskip
\end{minipage}
\vspace{-0.3cm}
\caption{FAP expression differences in real and virtual images.}
\label{FAP}
\end{figure}

We notice that in several cases FAP (purple connective tissue in FAP-CK images) expression is different between the real and virtual images images (Fig. \ref{FAP}). The localisation of FAP is generally successful however the patterns and the amounts are not always matching. One explanation for this effect can be that FAP is associated with tumor growth and increased angiogenesis rather than with anatomical or phenotypic features \cite{24}. If these functional features do not elicit a change visible in the input Ki67-CD8 staining, the model cannot correctly learn the mapping. The model success in localizing FAP, even if not flawlessly, suggests there are visual features in the input images that indicate the expression of FAP in the tissue. Identifying these visual features might lead to better FAP stain virtualisation as well as to new insights into tumor microenvironment anatomies. It is worthwhile to note that slide staining is a complex process with many variable and it is not uncommon to see variability in images of slides from the same tissue block stained with the same staining protocol due to variations in the tissue preparation, staining or imaging processes. For this reason, even with the visible difference in FAP expression in our virtual slides compared to the real stained slides, the results could still prove practically useful in some cases. Obtaining a virtual image which is slightly different from the real one should be acceptable as long as the pathologic interpretations of both of them are similar.

We validate our results on a dataset of 10 testing paired images of slides from the same tissue block using an automatic algorithm for CK\textsuperscript{+} and FAP cells detection. The algorithm was developed and validated using real FAP-CK images. The results include CK\textsuperscript{+} cell densities and FAP densities in real and virtual whole slide images. We verify that the difference between results on real and virtual stained slides is not one sided. This implies that our mapping algorithm does not consistently over-generate or under-generate CK or FAP.

\begin{figure}[t]

  \centering
  \centerline{\includegraphics[width=0.6\linewidth]{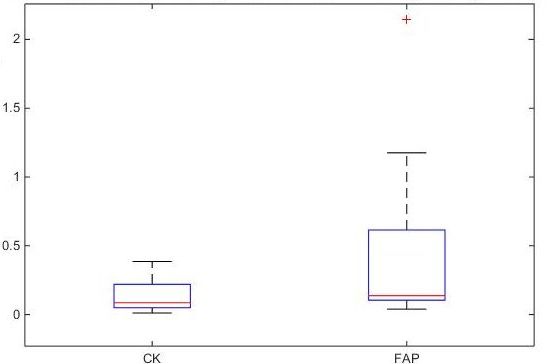}}
  
\caption{Boxplot representation of the absolute relative difference between real and corresponding virtual slides for CK\textsuperscript{+} cell  densities(left side) and FAP cell densities (right side).}
\label{cell_density_boxplot}
\end{figure}

In order to visualize the difference between these densities we compute the absolute relative difference between the results obtained in the real and virtual slides (Fig. \ref{cell_density_boxplot}). Analysis of the results shows a median absolute relative difference of 8\% with 0.016 variance between CK densities in real and virtual slides. This was also confirmed by our expert pathologist who evaluated real and virtual paired slides and reported high correlation in CK expression. For FAP, we report a median of 14\% with a variance of 0.466, reflecting a substantially higher variability than for CK. Our expert pathologist also confirmed this observation and mentioned that FAP features are completely not visible for pathologists in Ki67-CD8 staining. This observation is very interesting for our research and allows us to discover the limitations of simulation methods when biological constraints are present.

\section{Conclusions}
We propose to use an unsupervised deep learning method based on CycleGAN in order to virtually generate FAP-CK from Ki67-CD8 tissue stained images. Instance normalization used with the CycleGAN architecture helps the network learn a more realistic mapping between the stainings but introduces a tiling artifact in the merged testing set whole slide images. We significantly reduce this artifact by using a new inference approach based on overlapping tiles to create smooth transitions in the instance normalization layers. We validate our method using a test dataset of Ki67-CD8 input images for which a real FAP-CK images from the same tissue blocks were generated.

In the next steps, we plan to use additional input stainings and compare the results to our current results in order to define biological constraints to the success of stain virtualization. We also plan to replace the instance normalization module in the CycleGAN with an improved normalization element that reduces the titling effect during inference.

\bibliographystyle{IEEEbib}
{\footnotesize\bibliography{refs}}

\end{document}